\documentclass[sn-mathphys,Numbered]{sn-jnl}

\usepackage{graphicx}%
\usepackage{multirow}%
\usepackage{amsmath,amssymb,amsfonts}%
\usepackage{amsthm}%
\usepackage{mathrsfs}%
\usepackage[title]{appendix}%
\usepackage{xcolor}%
\usepackage{textcomp}%
\usepackage{manyfoot}%
\usepackage{booktabs}%
\usepackage{algorithm}%
\usepackage{algorithmicx}%
\usepackage{algpseudocode}%
\usepackage{listings}%
\usepackage{lettrine}
\usepackage{amsmath}
\usepackage{changes}

\theoremstyle{thmstyleone}%
\theoremstyle{thmstyletwo}%

\theoremstyle{thmstylethree}%

\begin{document}

\definechangesauthor[color=blue]{R1}
\definechangesauthor[color=purple]{R2}
\definechangesauthor[color=orange]{R3}

\title[Article Title]{SuPRA: Surgical Phase Recognition and Anticipation for Intra-Operative Planning}


\author*[]{\fnm{Maxence} \sur{Boels}*}\email{maxence.boels@kcl.ac.uk}
\author[]{\fnm{Yang} \sur{Liu}}
\author[]{\fnm{Prokar} \sur{Dasgupta}}

\author{\fnm{Alejandro} \sur{Granados}}
\equalcont{These authors contributed equally to this work.}

\author{\fnm{Sebastien} \sur{Ourselin}}
\equalcont{These authors contributed equally to this work.}

\affil[]{\orgdiv{Surgical and Interventional Engineering}, \orgname{King's College London}, \orgaddress{\city{London}, \country{UK}}}

\abstract{
Intra-operative recognition of surgical phases holds significant potential for enhancing real-time contextual awareness in the operating room. However, we argue that online recognition, while beneficial, primarily lends itself to post-operative video analysis due to its limited direct impact on the actual surgical decisions and actions during ongoing procedures. In contrast, we contend that the prediction and anticipation of surgical phases are inherently more valuable for intra-operative assistance, as they can meaningfully influence a surgeon's immediate and long-term planning by providing foresight into future steps.
To address this gap, we propose a dual approach that simultaneously recognises the current surgical phase and predicts upcoming ones, thus offering comprehensive intra-operative assistance and guidance on the expected remaining workflow.
Our novel method, Surgical Phase Recognition and Anticipation (SuPRA), leverages past and current information for accurate intra-operative phase recognition while using future segments for phase prediction. This unified approach challenges conventional frameworks that treat these objectives separately. We have validated SuPRA on two reputed datasets, Cholec80 and AutoLaparo21, where it demonstrated state-of-the-art performance with recognition accuracies of 91.8\% and 79.3\%, respectively.
Additionally, we introduce and evaluate our model using new segment-level evaluation metrics, namely Edit and F1 Overlap scores, for a more temporal assessment of segment classification.
In conclusion, SuPRA presents a new multi-task approach that paves the way for improved intra-operative assistance through surgical phase recognition and prediction of future events.
}


\maketitle

\section{Introduction}\label{sec1}

\lettrine[lraise=0.15, nindent=0em, slope=-.5em]{I}{ntra-Operative Surgical phase recognition} faces the intricate challenge of classifying frames in real-time.
Traditional methodologies offer recognition of events such as phases, steps, or action triplets, overlooking future information, they fall short in providing anticipative guidance to surgeons.
Existing approaches in the field have succeeded in mitigating some of those gaps by predicting future information, such as the next phase using statistical methods \cite{Franke_dtw_phase_ant}, and the remaining time until the occurrence of phases or instruments \cite{YUAN2022102611, rivoir20rethinking}.
Meanwhile, recent efforts have been oriented on leveraging temporal relations between observed frames \cite{DBLP:conf/miccai/CzempielPOKBN21, gao2021transsvnet} and under long temporal contexts \cite{Lovit2023, SKiT2023}. However, these intra-operative methods are predominantly built for recognition tasks and do not provide any prediction about future events that could guide surgeons through the surgical workflow.

In this work, we introduce a multi-task framework designed for joint surgical phase recognition and anticipation, that provides intra-operative guidance in the operating room in addition to post-operative video analysis.
It's important to delineate our use of the terms ``anticipation" and ``prediction", which we employ interchangeably, both referring to the classification of future events.
We use a recent state-of-the-art architecture called SKiT \cite{SKiT2023} to compress a long video into key features and decode those features into the current and future surgical phases along with their durations. Through this unified mutli-task approach, we aim to advance the state-of-the-art in surgical phase recognition, while also facilitating better preparation the upcoming surgical workflow through the prediction of the next phases.

Our approach centers on an encoder-decoder network called SuPRA Transformer (SuPRA), which is jointly trained to recognise current phases and predict upcoming ones. This novel architecture seeks to depart from conventional approaches that treat recognition and prediction as separate tasks. 
The primary contributions of this work are:
\begin{enumerate}
    \item a unified architecture designed for joint surgical phase recognition and anticipation with their segments duration,
    \item a rigorous evaluation on benchmark datasets, namely Cholec80 and AutoLaparo21, demonstrating state-of-the-art performance, and
    \item the introduction and use of new segment-level evaluation metrics for phases, i.e. Edit and F1 Overlap scores, for a temporal evaluation of the classified frames.
\end{enumerate}

\begin{figure}[t]%
\centering
\includegraphics[width=1.0\textwidth]{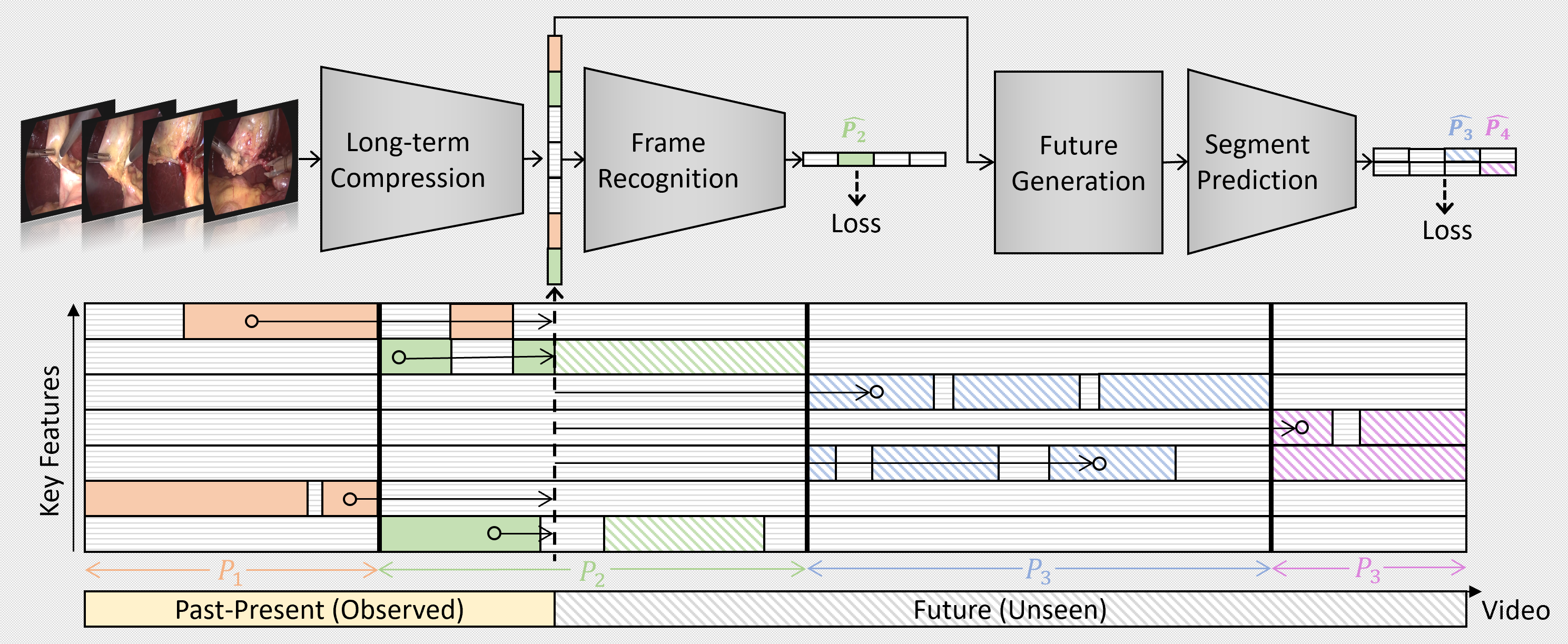}
\caption{Overview of our SuPRA model for joint phase recognition and prediction applied to surgical video analysis. \textit{Top:} Raw video frames undergo \textbf{Long-term Compression} to extract \textbf{Key-Features}. These are then classified into phases with the \textbf{Frame Recognition} module. The \textbf{Future Generation} module decodes those features into the future embeddings which are then classified by the \textbf{Segment Prediction} module, yielding the predicted upcoming phases. \textit{Bottom:} Video's temporal progression (x-axis) against its key-feature dimensions (y-axis). Solid vertical lines delimit individual phase segments, while the striped line indicates the current time \( t \). The filled color bars represent frames with observed key-features, while striped ones are yet to be observed. The left-to-right arrows represent the max-pooling operation on the compressed features while the right-to-left arrows represent the generated future key-features from the \textbf{Future Generation} module.}
\label{fig1}
\end{figure}

\section{Related Work}\label{sec2}

\noindent{\textbf{Surgical Phase Recognition.}} Early work in surgical phase recognition used Dynamic Time Warping, 
Hidden Markov Models, 
and statistical approaches.
With the advent of Convolutional Neural Networks (ConvNets) 
and the use of Backpropagation,
end-to-end spatial representations could be learned from convnets instead of using handcrafted feature detectors and extractors. 
Recurrent Neural Networks (RNNs), such as Long Short-Term Memory 
networks, were applied to model temporality between extracted spatial features thereafter \cite{DBLP:phd/hal/Twinanda17}. However, their ability to retain memory for long sequences is limited due to their vanishing gradients problem.
Temporal ConvNets (TCNs) \cite{DBLP:conf/eccv/LeaVRH16, DBLP:conf/cvpr/FarhaG19} were later used in TeCNO \cite{DBLP:conf/miccai/CzempielPKSFKN20} and Trans-SVNet (TSVN) \cite{gao2021transsvnet} to increase the field of view using convolutions. However, TCNs require a lot of memory, are limited to short videos, and the increased receptive field obtained through dilation can result in a loss of fine-grained details.
More recently, Self-Attention and Cross-Attention \cite{DBLP:conf/nips/VaswaniSPUJGKP17} brought a new paradigm to learn long temporal dependencies and quickly replaced RNNs in Natural Language Processing. Although Transformers can capture long temporal dependencies, their adoption for video recognition tasks remains limited due to the high memory requirements from the quadratic computational complexity resulting from self-attention and the number of pixels per video.
Previous works have successfully used Transformers to extract dense information between frames with short-term attention windows for multiscale temporal aggregation \cite{Lovit2023} or in combination with max-pooling to compress long-term context \cite{SKiT2023}.

\noindent{\textbf{Surgical Workflow Prediction.}}
The ability to anticipate e.g. predict the next gestures, actions, steps, or phases, is crucial in many applications, especially where autonomous systems could generate sequential planning and controls. 
In computer vision, recent works have shown that Transformers are highly effective in modeling spatiotemporal data for action anticipation \cite{girdhar2021anticipative, roy2022predicting}. The Anticipative Video Transformer (AVT) \cite{girdhar2021anticipative} is a state-of-the-art model that employs a Vision Transformer (ViT) \cite{dosovitskiy2020image} as its backbone and a causal transformer decoder to predict future features. Most popularised approaches in generative large language models use next-token prediction as a self-supervised training process to compress and decode long sequences of text in an auto-regressive manner. 
In the context of action anticipation, for instance, auto-regressive encoder-decoder models aim to generate future actions given a sequence of past frames \cite{zhang2021weaklysupervised}.
Conversely, parallel decoding has also been proposed to generate sequences of action segments characterised with long-term context that result in increased speed and consistency \cite{wang2023memoryandanticipation, abdelsalam2023gepsan} and improved action segmentation \cite{bahrami2023much}.

Anticipative approaches have been reported in the literature for surgical workflow analysis. These approaches used regression to predict lower-dimensional features such as the next phase using DTW \cite{Franke_dtw_phase_ant}, and the remaining time until the occurrence of phases or instruments \cite{YUAN2022102611, rivoir20rethinking}. TSVN proposed to jointly recognise the current phase with its remaining time duration until the next phase or instrument.
Although the concept of remaining time prediction for phase and instrument prediction were introduced as regression tasks, their approach didn't directly anticipate longer sequences of segments and their duration, which could be valuable for long-term planning.

\begin{figure}[h]%
\centering
\includegraphics[width=0.98\textwidth]{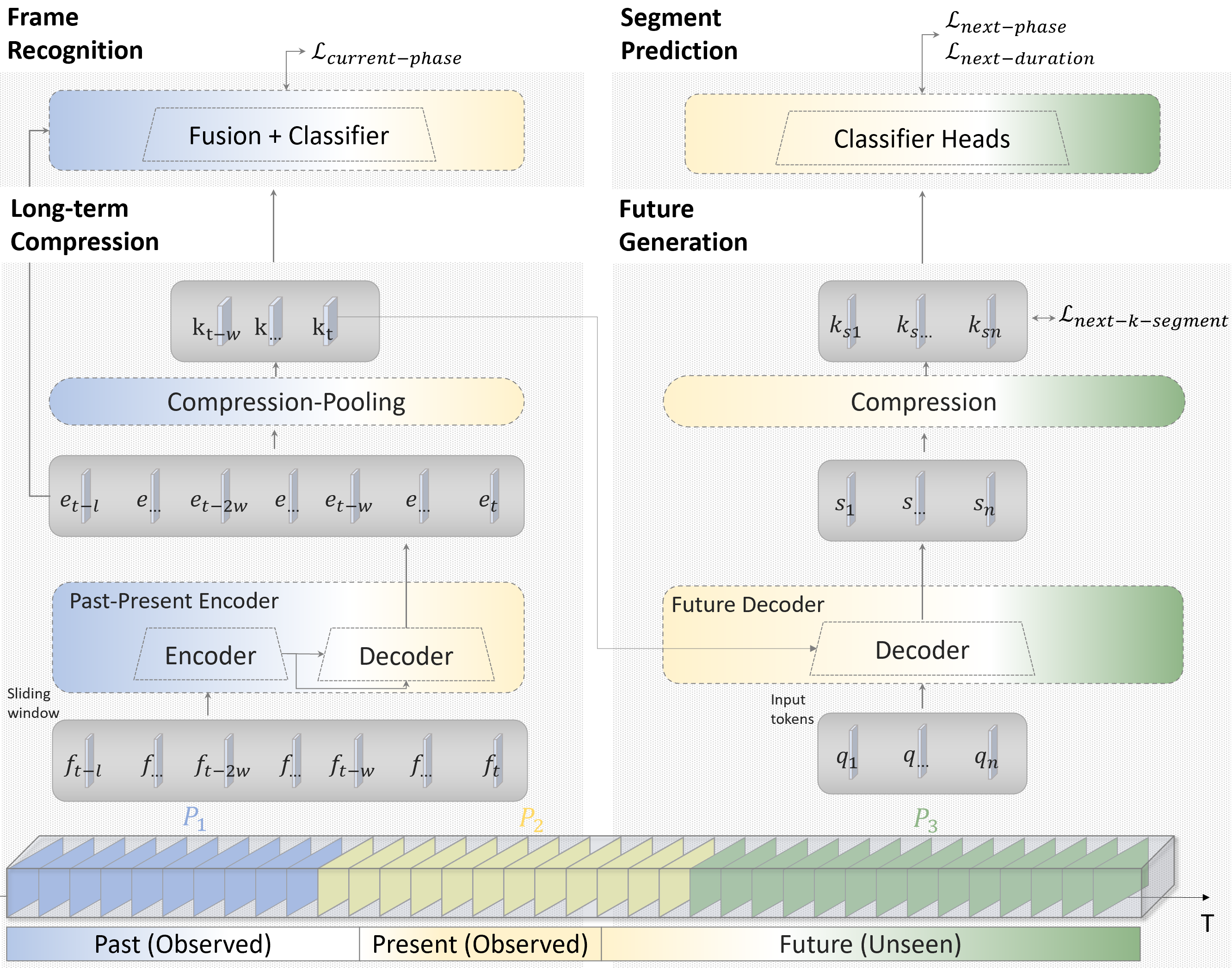}
\caption{SuPRA Transformer model. We use frame embeddings, $F = {\{f_1, ..., f_T\}}$ extracted with a ViT backbone  (omitted for simplicity) for all frames in a video $V = {\{x_1, ..., x_T\}}$. We first define an input clip as $c_t = {\{f_{t-l}, ..., f_t}\}$, where $l$ is the clip length. We then divide the clip into $l/w=n$ non-overlapping windows and define $w_t = {\{f_{t-w}, ..., f_t}\}$ as our input for the \textbf{Past-Present Encoder} to capture temporal patterns within this short clip. The \textbf{Sliding Window Attention} of length \( w \) ensures that all frames undergo Self-Attention. These frame embeddings are further refined using \textbf{Cross Attention} within the \textbf{Past-Present Decoder}, encapsulated by the encoder-decoder module. For a concise representation of salient video features, \textbf{Long-term Compression} implements compression-pooling on all the encoded embeddings, yielding \( w \) keys \( k_{t-w}, \ldots, k_{t} \). In parallel, the \textbf{Future Generation} module exploits the \textbf{Future Decoder} to transform \( n \) segment queries \( q_1, q_2, \ldots, q_n \) into \( n \) future decoded segment \( s_1, \ldots s_n \), thereby forecasting unseen future segments i.e. phases. These generated segments are then compressed to \( n \) key-segment \( k_{s1}, \ldots, k_{sn} \), used to guide the decoder with a supervisory next-key-segment loss, \( L_{\text{next-k-segment}} \). The \textbf{Segment Prediction} module translates decoded key-segments into subsequent phases, denoted \( \hat{y}_{1}, \ldots, \hat{y}_{n} \), along with their durations \( \hat{d}_{1}, \ldots, \hat{d}_{n} \), optimising the predictions via multiple losses—\( L_{\text{next-phase}} \) and \( L_{\text{next-duration}} \)—through classifier heads. Lastly, the \textbf{Frame Recognition} integrates the processed data for final classification \( \hat{y}_{t-w}, \ldots, \hat{y}_{t} \).}\label{fig2}
\end{figure}

\section{Methods}\label{sec3}
We train SuPRA for both phase recognition and anticipation, while predicting the duration of subsequent phases. This multi-task approach allows for the simultaneous execution of two fundamental tasks intra-operatively.

\subsection{Spatial Feature Extractor}
The spatial feature extractor backbone (omitted for simplicity in Figure \ref{fig1}) is designed to address the problem of ambiguous frames that arises when using single-frame inputs for frame classification. Similar to \cite{Lovit2023}, we use multiple frames during training to learn short spatiotemporal representations. The backbone of our model is trained following AVT \cite{girdhar2021anticipative} to encode each frame $x_t$, observed at time $t$, of a video $V={\{x_1, ..., x_T}\}$ into a compressed representation $F={\{f_1, ..., f_T}\}$, where $f_t = B(x_t)$.

\subsection{Long-Term Compression}
To classify frame embeddings $F$ extracted by the backbone $B$, we define a long video clip as $c_t = {\{f_{t-l}, ..., f_t}\}$, where $l$ is the video clip length. 

\noindent\textbf{Past-Present Encoder}. We use a window of length $w$ to perform sliding window self-attention on the long video clip $c_t$ with stride $w$. Our past-present encoder-decoder, $ED_{PP}$ takes an input sequence $c_t$ of length $l$ and outputs a sequence of the same length, $E_t = {\{e_{t-l}, ..., e_t}\}$.

\noindent\textbf{Compression-Pooling}. Short temporal sliding windows with no overlap may not capture long-term dependencies between frames, and overcome ambiguous video segments. To address the need for long-term dependencies, we propose to aggregate the long sequence of encoded short-temporal spatial features $E_t = {\{e_{t-l}, ..., e_t}\}$ with max-pooling and compress their representation in a lower-dimensional space, $K_t = {\{k_{t-w}, ..., k_t}\}$ to capture the most relevant observed encoded features as proposed in \cite{SKiT2023}. This process, known as \textit{key-pooling}, aggregates all the compressed frames by selecting the maximum value in each dimension up to the $i^{th}$ frame, which retrieves the most relevant information observed so far in the video, yielding in a single feature vector of size $d$.

\subsection{Future Generation}
This module has a dual objective, it primarily aims to provide future information, but also attempts to learn better representations about the present for the recognition task. We therefore explore the effect of anticipation on the recognition task and evaluate the anticipation capabilities of our network.

\noindent\textbf{Future Decoder}. Our proposed approach uses a \textit{Future Decoder}, $Dec_{F}$, which takes as input the compressed maxpooled features $K_t = {\{k_{t-w}, ..., k_t}\}$ and decodes input queries, $Q = {\{q_1, q_2, \ldots, q_n}\}$ into \( n \) future decoded segments $S = {\{s_1, s_2, \ldots, s_n}\}$.

\noindent\textbf{Compression}. After decoding the \( n \) input queries with the long-term compressed key-freatures, $K_t = {\{k_{t-w}, ..., k_t}\}$, we extract the key information from the decoded future segments using a linear compression layer, $Comp_{FS}$.

\subsection{Frame Recognition}

\noindent\textbf{Fusion}. We fuse $K_t$ and the last $w$ frames in $E_t$ by summing and using a skip connection with a fully connected layer, following \cite{SKiT2023}.

\noindent\textbf{Classifier}. The resulting feature vector is then passed through a classification layer to predict class probabilities for ${{x_{t-w}, ..., x_t}}$.The final prediction $\hat{y_t}$ for frame $x_t$, takes the last frame's class probabilities, $\hat{p_t}$, which is trained using a consistency cross-entropy loss function proposed in \cite{DBLP:conf/cvpr/FarhaG19} to improve temporal consistency and helps smooth the predictions over time.

\subsection{Segment Prediction}

\noindent\textbf{Classifier Heads}. We use two classification heads to jointly predict the \textit{next phases} and their \textit{duration}. Note, that we can also predict the next key-features for more supervision. Optionally, we predict the remaining time in the current phase along with its phase label $\hat{y_t}$, for frame $x_t$.

\subsection{Training Objectives}

The total loss function \(\mathcal{L}_{\text{total}}\), is defined as:

\[
\mathcal{L}_{\text{total}} = \mathcal{L}_{\text{current-phase}} + \mathcal{L}_{\text{next-phase}} + \mathcal{L}_{\text{next-duration}} (+ \mathcal{L}_{\text{next-k-features}})
\]

where each component is defined as:

\textbf{Current and Next Phase Loss (CE):} \(\mathcal{L}_{\text{current-phase}}\) and \(\mathcal{L}_{\text{next-phase}}\) leverage cross-entropy loss to enhance the accuracy of class predictions for the current and subsequent phases.

\textbf{Next Duration Loss (MSE):} \(\mathcal{L}_{\text{next-duration}}\) employs Mean Squared Error (MSE) to provide precise estimations of the duration of upcoming classes.

\textbf{Next Key Features Loss (MSE):} Similarly, \(\mathcal{L}_{\text{next-k-features}}\) utilises MSE, focusing on the accurate reconstruction of key features for forthcoming phases. These features are extracted via a Long-term Compression module.

\subsection{Segment-based Evaluation Metrics}
Traditional metrics like precision, recall, and the Jaccard index, while useful for assessing overall accuracy, often fail to capture crucial temporal dynamics. They focus primarily on the correctness of each individual frame's classification, without adequately considering the continuity and progression of actions over time. This limitation is addressed by Edit Score and F1 Overlap, which offer a refined evaluation of temporal action segmentation tasks \cite{ding2022temporal}.

\textbf{Edit Score} \cite{lea16segmental} evaluates the alignment of predicted segment boundaries with ground truth, factoring in the sequence and order of actions. This metric is particularly adept at handling over-segmentation errors, where a continuous action is erroneously divided into multiple segments. Unlike precision and recall, the Edit Score is less sensitive to such over-segmentation \cite{Ishikawa_2021_WACV}, as it can correct a series of fragmented segments with a single edit operation. This tolerance makes it a suited metric for assessing the temporal structure and flow of activities in videos, an aspect overlooked by metrics such as Jaccard, precision, and recall.

\textbf{F1 Overlap} \cite{DBLP:conf/eccv/LeaVRH16} extends the concept of precision and recall to the temporal domain, evaluating the overlap between predicted and ground truth segments at various thresholds (e.g., $10$\%, $25$\%, $50$\%). F1 Overlap provides a flexible yet precise measure of how accurately a model captures both the duration and timing of actions. Its sensitivity to the chosen threshold allows for different levels of strictness in evaluation.

\textbf{Frame-Level Accuracy}, which measures the proportion of correctly classified frames, is crucial for evaluating frame misclassification errors. However, it does not address over-segmentation issues, which relate to the temporal arrangement of actions rather than the correctness of frame classification.

In summary, while precision, recall, and the Jaccard index offer a foundational understanding of model performance, the Edit Score and F1 Overlap provide a complementary view that is essential for evaluating temporal segmentations.

\subsection{Implementation Details}
Our experiments were conducted on a single NVIDIA Tesla V100 GPU. We used a 12-head, 12-layer Transformer encoder as our spatial feature extractor, based on the ViT-B/16 architecture following LoViT \cite{Lovit2023}. This model was pre-trained on ImageNet 1K (IN1k) \cite{deng2009imagenet} and produced 768-d representations, with an input image size of 248$\times$248 pixels.
For training the spatial feature extractor, we used stochastic gradient descent with momentum for 35 epochs, with a 5-epoch warm-up period 
and a 30-epoch cosine annealed decay. 
We used a batch size of 16 and a learning rate of 0.1, which was multiplied by 0.1 at the 20th and 30th epochs. We set the weight decay to 1e-4 and the momentum to 0.9.
For the past-present encoder, we used an input clip length of 3000 frames with a sliding window of length 20, generating 512-d feature vectors. Key-pooled feature dimensions are 64-d and 32-d on Cholec80 and AutoLaparo21, respectively. The temporal modules underwent training for 40 epochs using SGD and momentum with a learning rate of 3e-4, weight decay of 1e-5, a 5 epoch warm-up period, and a 35 epoch cosine annealed decay, with a batch size of 8.

\section{Experiments and Results}
\subsection{Experimental Setup}

\noindent\textbf{Datasets.} We evaluate our work on two publicly available surgical phase recognition datasets, namely \textit{Cholec80} (C80) \cite{endonet} and \textit{AutoLaparo21} (AL21) \cite{DBLP:conf/miccai/WangLLZCDL22}.
For C80, we keep the same video splits as in \cite{gao2021transsvnet, Lovit2023} with the first 40 videos for training and the other half for evaluation. C80 has 7 classes i.e. surgical phases. We do not use the tool presence annotations and only predict the phases.
In AL21, we follow the same splits as in \cite{DBLP:conf/miccai/WangLLZCDL22, Lovit2023} where 21 videos are divided into 10, 4, and 7 videos for training, validation, and testing, respectively. Both datasets were sampled at 1 frame per second (fps) following previous works \cite{gao2021transsvnet, Lovit2023}.

\noindent\textbf{Baseline Selection and Comparison.} In this study, we have carefully selected our baselines to align with our research methodology and objectives. The models chosen for comparison are SKiT \cite{SKiT2023} and Trans-SVNet (TSVN) \cite{gao2021transsvnet}, which share similarities with our approach. For \textit{phase recognition} with anticipation, we change the regression task in TSVN \cite{gao2021transsvnet} to a classification task. For \textit{phase prediction}, we keep TSVN for its dual tasks, except that instead of solely predicting the remaining time until the occurrence of instruments or phases, we modify the target to become the next phase. Note that we didn't select \cite{rivoir20rethinking, YUAN2022102611} since they both evaluate anticipation as a regression task and \cite{YUAN2022102611} leverages other tasks for supervision i.e. segmentation and detection.

\subsection{Phase Recognition}
We evaluate our SuPRA architecture on \textit{phase recognition} and study the effect of anticipation on the recognition task. We compare our results, with different supervision tasks i.e. phase recognition with and without anticipation, on both datasets in Table \ref{tab1}.

\begin{table}[h]
\caption{Comparative analysis of state-of-the-art methods for phase recognition on Cholec80 and AutoLaparo21 datasets. The results underscore the robustness of our method, exhibiting its competencies on both frame- and segment-level metrics, while offering an additional task. While SKiT achieves a marginally better accuracy on Cholec80, SuPRA is on par with the Edit score and F1-score metrics. Our method, with anticipation enabled, also achieves superior performance on AutoLaparo21 for all metrics.}\label{tab1}
\begin{tabular*}{\textwidth}{@{\extracolsep\fill}l|c|ccccc|ccccc}
\toprule
& & \multicolumn{5}{@{}c@{}}{Cholec80} & \multicolumn{5}{@{}c@{}}{AutoLaparo21} \\ \cmidrule{2-11}
Methods & Ant. & \multicolumn{1}{c}{Acc.} & \multicolumn{1}{c}{Edit} & \multicolumn{3}{c}{F1\{10, 25, 50\}} & \multicolumn{1}{c}{Acc.} & \multicolumn{1}{c}{Edit} & \multicolumn{3}{c}{F1\{10, 25, 50\}} \\
\midrule
TSVN* \cite{gao2021transsvnet}  & $\times$ & $88.87$ & $12.69$ & $21.62$ & $20.86$ & ${18.04}$ & $\mathbf{-}$ & $\mathbf{-}$ & $\mathbf{-}$ & $\mathbf{-}$ & $-$ \\
\midrule
TSVN* \cite{gao2021transsvnet}  & $\checkmark$ & $88.75$ & $12.62$ & $21.29$ & $20.72$ & ${17.55}$ & $\mathbf{-}$ & $\mathbf{-}$ & $\mathbf{-}$ & $\mathbf{-}$ & $-$ \\
\midrule
SKiT \cite{SKiT2023}  & $\times$ & $\mathbf{92.4}$ & $21.9$ & $34.0$ & $33.4$ & $\mathbf{29.7}$ & $77.3$ & $6.6$ & $11.2$ & $9.7$ & $5.0$ \\
\botrule
SuPRA(Ours) & $\times$ & $91.8$ & $\mathbf{21.9}$ & $\mathbf{34.0}$ & $\mathbf{33.6}$ & $28.1$ & $77.0$ & $7.1$ & $11.6$ & $9.9$ & $5.5$ \\
\midrule
SuPRA(Ours) & $\checkmark$ & $91.3$ & $21.2$ & $33.1$ & $32.16$ & $27.2$ & $\mathbf{79.3}$ & $\mathbf{8.8}$ & $\mathbf{14.6}$ & $\mathbf{12.1}$ & $\mathbf{6.9}$ \\
\botrule
\end{tabular*}
\footnotetext{Note: * indicates methods we reproduced using the released weights on Cholec80.}
\end{table}

\subsection{Next Phase Prediction:}
For this task, we maintain TSVN as a baseline due to its dual-task capability.
We change the remaining time target to predict the next phase and report our results in Table \ref{tab2}.

\begin{figure}[t]
\centering
\includegraphics[width=1.\textwidth]{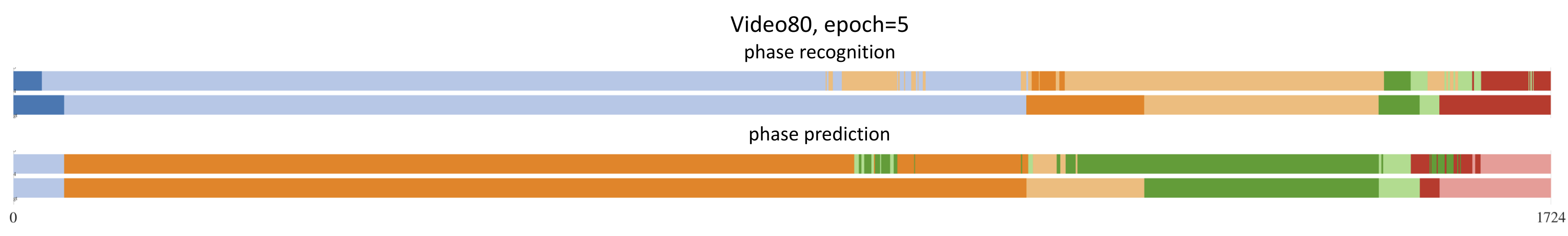}
\caption{Qualitative illustration of the phase recognition (top 2 rows) and next phase prediction (bottom 2 rows) tasks for video 80 from the Cholec80 dataset with online frame classification results (top row) and their annotations (bottom row). We add an ``end" class in our evaluation to replace the first class (pink segment).}\label{fig3}
\end{figure}

\begin{table}[]
\caption{Comparative analysis of state-of-the-art methods for next phase prediction on Cholec80 and AutoLaparo21 datasets. The performance of each method is evaluated based on Accuracy (Acc.), Edit score, and F1 Overlap metrics at different thresholds (10\%, 25\%, 50\%).}\label{tab2}
\begin{tabular*}{\textwidth}{@{\extracolsep\fill}l|ccccc|ccccc}
\toprule
& \multicolumn{5}{@{}c@{}}{Cholec80} & \multicolumn{5}{@{}c@{}}{AutoLaparo21} \\ \cmidrule{1-11}
Methods & \multicolumn{1}{c}{Acc.} & \multicolumn{1}{c}{Edit} & \multicolumn{3}{c}{F1\{10, 25, 50\}} & \multicolumn{1}{c}{Acc.} & \multicolumn{1}{c}{Edit} & \multicolumn{3}{c}{F1\{10, 25, 50\}} \\
\midrule
TSVN \cite{gao2021transsvnet} & $77.71$ & $11.58$ & $17.62$ & $16.01$ & $13.60$ & $\mathbf{-}$ & $\mathbf{-}$ & $\mathbf{-}$ & $-$ & $-$ \\
\midrule
SKiT \cite{SKiT2023}  & $\mathbf{83.7}$ & $15.3$ & $23.4$ & $22.3$ & ${18.5}$ & $60.2$ & $5.3$ & $6.8$ & $5.5$ & $2.6$ \\
\midrule
SuPRA (Ours)     & $83.3$ & $\mathbf{16.4}$ & $\mathbf{24.8}$ & $\mathbf{23.2}$ & $\mathbf{20.1}$ & $\mathbf{66.1}$ & $\mathbf{6.9}$ & $\mathbf{9.6}$ & $\mathbf{7.4}$ & $\mathbf{4.6}$ \\
\botrule
\end{tabular*}
\footnotetext{Note: we trained and evaluate both methods directly on the next phase prediction task with an additional ``end" class. Values in bold indicate the best performance for each metric.}
\end{table}

\subsection{Ablation Study and Analysis}
\noindent\textbf{Multiple segment prediction.} This experiment aims to evaluate how accurately one can predict multiple segments at once for workflow generation. As shown in Table \ref{tab3}, our model can predict the next 6th segment with up to $69.2$\% and $50.1$\% accuracy ($Acc._{\text{Pred}_{6}}$) on Cholec80 and AutoLapro21, while maintaining state-of-the-art performance on surgical phase recognition ($Acc._{\text{Rec}}$).

\begin{table}[h]
\caption{Comparison of recognition accuracy ($Acc._{\text{Rec}_{0}}$) and prediction accuracies ($Acc._{\text{Pred}_{1}}$-$Acc._{\text{Pred}_{4}}$)  with varying number of anticipated segments. We increase the number of predicted segments from 0 to 4 and observe how much multiple predictions can affect the accuracies. The table illustrates the model's ability to consistently recognise the current phase with high accuracy while also predicting forthcoming segments effectively.}\label{tab3}
\begin{tabular*}{\textwidth}{@{\extracolsep\fill}c|c|ccccc}
\toprule%
Predicted & \multicolumn{5}{@{}c@{}}{Cholec80}\\
Segments & $Acc._{\text{Rec}}$ &  $Acc._{\text{Pred}_{1}}$ &  $Acc._{\text{Pred}_{2}}$ &  $Acc._{\text{Pred}_{3}}$ &  $Acc._{\text{Pred}_{4}}$ \\
\midrule
0 & $\mathbf{91.8}$ & - & - & - & - \\ 
1 & $91.3$ & $\mathbf{83.3}$ & - & - & - \\
2 & $91.0$ & $81.4$ & $76.1$ & - & -  \\
4 & $91.3$ & $82.3$ & $75.8$ & $77.0$ & $56.8$  \\
\botrule
Predicted & \multicolumn{5}{@{}c@{}}{AutoLaparo21}\\
Segments & $Acc._{\text{Rec}}$ &  $Acc._{\text{Pred}_{1}}$ &  $Acc._{\text{Pred}_{2}}$ &  $Acc._{\text{Pred}_{3}}$ &  $Acc._{\text{Pred}_{4}}$ \\
\midrule
0 & $77.0$ & - & - & - & - \\
1 & $\mathbf{79.3}$ & $\mathbf{66.1}$ & - & - & - \\
2 & $78.1$ & $61.7$ & $57.7$ & - & - \\
4 & $77.4$ & $59.4$ & $48.1$ & $33.5$ & $32.3$ \\
\botrule
\end{tabular*}
\end{table}

\noindent\textbf{Does Anticipation Improves Recognition?} Our analysis, based on Tables \ref{tab1}, \ref{tab2}, and \ref{tab3}, unveils an interplay between the value of future information for phase recognition. Specifically, we observe consistent improvement in recognition accuracy, Edit, and F1 scores when predicting future segments on AL21 whereas no improvement is observed for recognition metrics when predicting future segments on Cholec80. Therefore, depending on the dataset, future information can effectively enhance the model's capability for surgical phase recognition. Additionally, we note a decreasing trend in prediction accuracies ($Acc._{\text{Pred}_{1}}$-$Acc._{\text{Pred}_{6}}$) as the anticipation is extended, likely due to the increased uncertainty and complexity associated with forecasting more distant future segments.

\section{Discussion}\label{sec:discussion}

Our work highlights the possibility for improving and augmenting current surgical phase recognition models with anticipative tasks such as next-phases prediction and their durations.

\noindent\textbf{Comparative Effectiveness for Anticipative Phase Recognition.} 
In evaluating the effectiveness of anticipative recognition for surgical phases, our analysis reveals insightful distinctions between datasets and models. On the Cholec80 dataset, SuPRA achieves a notable accuracy of 91.3\%, competing closely with the state-of-the-art SKiT model, which marginally leads with 92.4\% accuracy. However, we highlight that SuPRA matches SKiT in terms of Edit score (21.9\%) and F1 scores. For the AutoLaparo21 dataset, the segment-level anticipation of SuPRA achieves a peak accuracy of 79.3\%, which is higher than SKiT's 77.3\%. More importantly, SuPRA outperforms SKiT across all metrics in this dataset, reinforcing the model's robustness in handling more complex or varied surgical scenarios. The comparative analysis, as detailed in Table 1, underscores the robustness of SuPRA, exhibiting its competencies in both frame- and segment-level metrics. Notably, with anticipation enabled, SuPRA demonstrates even superior performance on the AutoLaparo21 dataset across all evaluated metrics.

\begin{figure}[t]%
\centering
\includegraphics[width=.8\textwidth]{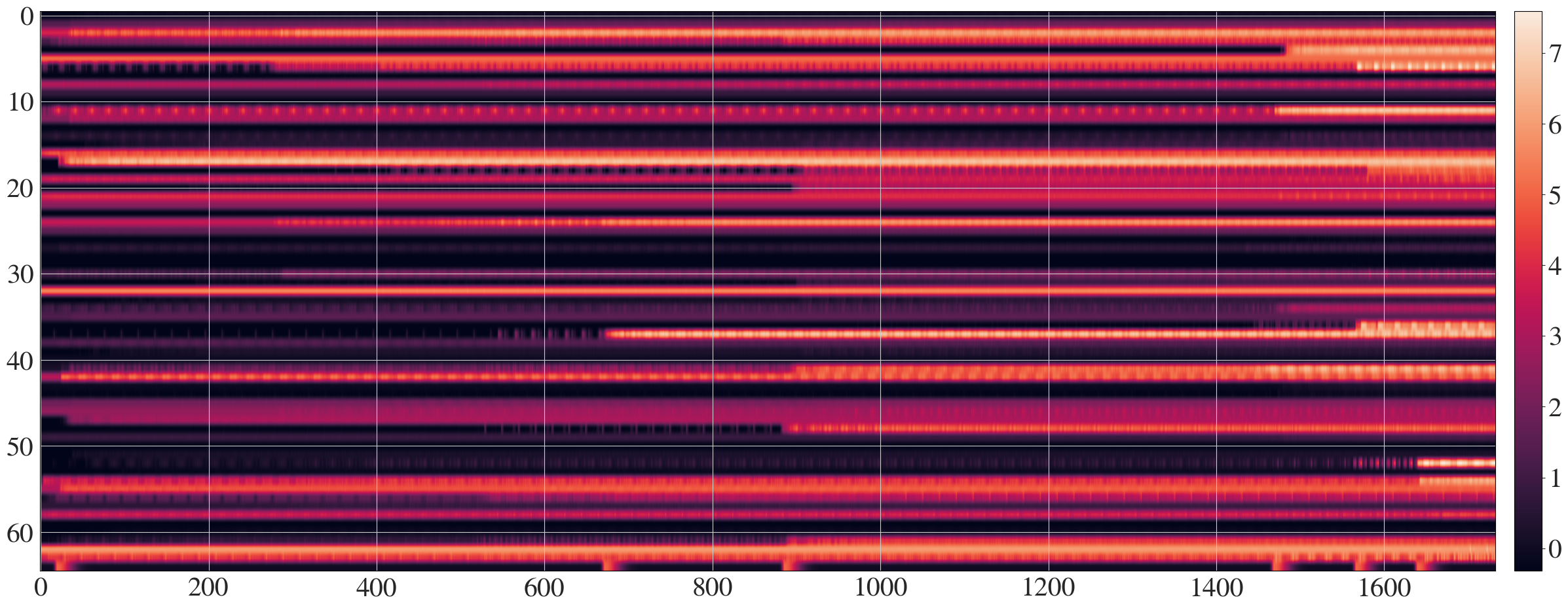}
\caption{Temporal Aggregation of Key-Features. The x-axis represents the sequential frames of the video, while the y-axis denotes the compressed key dimensions. For each frame, maxpooling is employed to aggregate the salient key-features, leading to a temporal accumulation of feature prominence as depicted from the increasing intensity variations. This methodology capitalises on the notion that a cumulative representation of dominant events and features over time enhances the comprehension of the current surgical phase. The goal is to leverage this aggregated information to anticipate and generate future compressed representations.}\label{fig3}
\end{figure}

\noindent\textbf{Comparative Effectiveness for Next Phase Prediction.}
Our comprehensive evaluation our SuPRA method against state-of-the-art alternatives TSVN and SKiT for surgical phase prediction on the Cholec80 and AutoLaparo21 datasets offers significant insights into the effectiveness of these approaches. On the Cholec80 dataset, SuPRA demonstrates a highly competitive performance with an accuracy of 83.3\%, closely trailing SKiT's 83.7\%. More notably, SuPRA establishes its superiority in the Edit score (16.4\%) and F1 scores across various thresholds (10, 25, 50), consistently outperforming other methods. This makes SuPRA a compelling choice for accurate next-phase anticipation.

For the AutoLaparo21 dataset, SuPRA's advantages become even more pronounced. Leading the field with an accuracy of 66.1\%, it also shows substantial improvements for the Edit (6.9\%) and F1 scores at all thresholds (9.6\%, 7.4\%, 4.6\%). This performance is a testament to SuPRA's adaptability and strong capability in both surgical procedures.

Overall, our results, as detailed in Table 2, underscore SuPRA's balanced and superior performance across almost all evaluated metrics on both datasets. This positions SuPRA as a robust and versatile tool for next-phase prediction in surgical procedures.

\noindent\textbf{Limitations and Future Directions.}
While our SuPRA model exhibits notable strengths, we acknowledge several limitations in our current approach. Firstly, a significant concern arises from the observed low Edit and F1 scores across all models, including SuPRA, as evident in our results. These low scores suggest a tendency for the model to oscillate in its predictions, leading to frequent changes in phase recognition and anticipation. The low scores might stem from several factors, including the inherent complexity and variability of surgical procedures, the coarse labelling process, or potential limitations in the current model architecture and training process.

In future work, we aim to further evaluate the predicted phase durations. This involves reconstructing future segments not only in the correct sequential order but also in their actual length. Such an approach could significantly enhance the precision and utility of surgical workflow generation. Looking ahead, there are other several promising directions for future research. One such direction involves testing our model on more intricate tasks that have less structured sequences, such as surgical steps or actions prediction, where the order of events is less predictable. Additionally, we are interested in exploring the potential of autoregressive decoding, similar to methods used in text generation, to assess its efficacy in intra-operative surgical planning.

\section{Conclusion}\label{sec:conclusion}

We introduced SuPRA Transformer, a novel architecture that successfully integrates past, present and future information for surgical phase recognition and anticipation. This unified model disrupts the traditional dichotomy between phase recognition and phase prediction, thus offering a more holistic solution.
Our method was rigorously evaluated on the Cholec80 and AutoLaparo21 datasets, where it exhibited state-of-the-art performance on frame- and segment-level metrics.
In summary, this research not only matches the state-of-the-art in surgical phase recognition but also opens the door for future work on surgical workflow generation for intra-operative guidance.

\bibliography{sn-article}

\end{document}